# A novel total variation model based on kernel functions and its application

Zhizheng Liang, Lei Zhang, Jin Liu, Yong Zhou, School of Computer Science and Technology, China University of Mining and Technology, Xu Zhou, 221116,China

E-mail: liang@cumt.edu.cn

**Abstract:** The total variation (TV) model and its related variants have already been proposed for image processing in previous literature. In this paper a novel total variation model based on kernel functions is proposed. In this novel model, we first map each pixel value of an image into a Hilbert space by using a nonlinear map, and then define a coupled image of an original image in order to construct a kernel function. Finally, the proposed model is solved in a kernel function space instead of in the projecting space from a nonlinear map. For the proposed model, we theoretically show under what conditions the mapping image is in the space of bounded variation when the original image is in the space of bounded variation. It is also found that the proposed model further extends the generalized TV model and the information from three different channels of color images can be fused by adopting various kernel functions. A series of experiments on some gray and color images are carried out to demonstrate the effectiveness of the proposed model.

Keywords: total variation; kernel functions; Hilbert space; image enhancement

## 1 Introduction

Over the past several decades partial differential equations (PDEs)[1-5] have been widely used in image processing such as image denoising, edge detection, image deconvolution and image inpainting because they provide more intuitive mathematical models. These models based on PDEs [6-10] have been established to modify an image or a surface by evolving their solutions. One type of methods used to obtain PDEs [11-15] is that an energy functional is first defined in terms of the expected properties of the output image and the evolution equation is then derived by computing the Euler-Lagrange equation of this energy functional. Among this category of methods, the typical and effective one was proposed by Rudin, Osher and Fatemi [4] in 1992. They defined a certain total variation as a penalty function in a minimization problem and this model is usually referred to as the ROF model. Introducing the total variation, however, leads to a non-smooth objective function. Subsequently, several effective schemes have been proposed to address the non-smooth problem [3]. Note also that the ROF model suffers from undesirable results for image processing such as the staircase effect and the loss of texture. Some improved ROF models [5-7] have also been proposed to alleviate this problem. For example, Song [5] proposed a generalized total variation model which effectively reduces the staircase effect in the TV model and still keeps sharp edges. Osher et al.[6] developed an iterative regularization method where the squared L2 norm in the fidelity term is replaced by the L1 norm. Esedoglu and Osher [7] introduced an anisotropic version of the ROF model to remain certain edge directions. In addition, Gilboa and Osher [16] also proposed a nonlocal total variation model that combines the idea of variational models and patch-based methods in nonlocal means. Overall, these methods continue to contribute to the development of PDE-based models in some sense.

Some of total variation models mentioned above involve solving nonlinear equations. The nonlinearities of equations may capture the complex structure of images and enable this category of approaches to achieve good performance in the general case. It is noted that kernel functions in kernel learning methods [18-19] are a class of nonlinear functions. These kernel functions can effectively deal with some nonlinear problems and have gained wide applications in many areas such as pattern recognition and computer vision. Inspired by the idea of kernel learning methods, we expect that introducing these kernel functions into previous total variation models will further improve the performance of these models since this increases the nonlinearity of these equations which changes the dynamic range of gray levels of images. To this end, in this paper we propose a new variation model based on kernel functions. In the proposed model, the original image space is first mapped into a high-dimensional feature space and a nonlinear equation in a feature space is achieved from the total variation model. Then a kernel equation is derived from a couple of equations and evolves in a kernel function space. Finally, by considering the characteristics of kernel functions and utilizing the coupled image, we can obtain the processed image in the image or kernel space. Note that kernel functions in the proposed model may serve as prior knowledge of images in some sense, and they also provide a strategy for fusing the information from three different channels of color images. In addition, we also carry out some experiments to show the effectiveness of the proposed model on gray-level and color images.

## 2 Generalized TV models

In this section, we briefly introduce the generalized TV model which is defined as the following optimization problem [5]:

$$\min J_p(I^a) = \frac{1}{p}\int_Q |\nabla I^a|^p dxdy + \frac{\lambda}{2}\int_Q |I^a - I_0^a|^2 dxdy, \tag{1}$$

where $Q$ is the domain of definition of the image, $I_0^a$ is the degraded image to restore, $\lambda$ is a tradeoff parameter, $|\nabla(I^a)| = \sqrt{\langle \partial_x(I^a), \partial_x(I^a) \rangle + \langle \partial_y(I^a), \partial_y(I^a) \rangle}$, $\nabla I^a = (\partial_x I^a, \partial_y I^a)$, and $p$ is a positive number. It is found that Eq. (1) degenerates into the classical ROF model if $p=1$. It is obvious that the processed image can be achieved by solving Eq.(1). Note that if the image to be processed is a gray-level image, $I^a$ contains a component; if the image is a color image, $I^a = (I_1^a, I_2^a, I_3^a)^T$ contains three components, where $I_1^a, I_2^a$, and $I_3^a$ denote red, green, and blue components of a color image respectively. The associated Euler-Lagrange equation of Eq.(1) is written as

$$-\nabla \cdot (|\nabla I^a|^{p-2} \nabla I^a) + \lambda(I^a - I_0^a) = 0. \tag{2}$$

$\nabla \cdot$ denotes the divergence operator and is to be taken componentwise. In fact, there are usually two schemes to solve Eq.(1). The one is to directly solve the Euler-Lagrange equation of Eq.(1) and the other is to apply the gradient decent method to solve Eq.(1). From Eq.(2), one may obtain a parabolic equation by introducing time as an evolving parameter, denoted by

$$\frac{\partial I^a}{\partial t} = \nabla \cdot (|\nabla I^a|^{p-2} \nabla I^a) - \lambda(I^a - I_0^a). \tag{3}$$

## 3 Generalized Kernel TV model
### 3.1 Kernel Methods

Assume that a nonlinear map $\phi$ is given, an input data space $\Re$ can be mapped into a feature space $F$ [19-20].

$$\phi : \Re \to F, I^a(x,y) \mapsto \phi(I^a(x,y)). \tag{4}$$

In terms of this nonlinear map, the pixel at coordinates $x$ and $y$ in an image space is mapped into a much higher dimensional vector in a feature space. Specifically, the value of the pixel $I^a(x,y)$ at coordinates $x$ and $y$ in a gray or color image is projected to $\phi(I^a(x,y))$ by using the nonlinear map $\phi$. Note that the feature space could have an arbitrarily large, possibly infinite, dimensionality and it is also difficult to give an explicit expression of this map $\phi$. However, by defining an inner product operation, one can obtain a kernel [19-20]. A kernel is a function $k$ that satisfies $k(I^a(x,y), I^b(x,y)) = <\phi(I^a(x,y)), \phi(I^b(x,y))>$, where $I^b(x,y)$ denotes another image. As a result, one can compute the inner product of two input vectors in a feature space without knowing the explicit mapping function. The two widely used kernel functions in machine learning are polynomial kernels and Gaussian kernels, denoted by $k(I^a(x,y), I^b(x,y)) = \langle I^a(x,y), I^b(x,y) \rangle^d$ and $\exp(-(|I^a(x,y) - I^b(x,y)|^2 / 2\delta))$, where $d$ and $\delta$ are two kernel parameters.

### 3.2 Generalized TV models based on kernel functions
### 3.2.1 The proposed model

Using the nonlinear map defined in Eq.(4), one can obtain the following equation from Eq.(3).

$$\frac{\partial \phi(I^a)}{\partial t} = \nabla \cdot (|\nabla \phi(I^a)|^{p-2} \nabla \phi(I^a)) - \lambda(\phi(I^a) - \phi(I_0^a)), \tag{5}$$

where $|\nabla \phi(I^a)| = \sqrt{\langle \partial_x \phi(I^a), \partial_x \phi(I^a) \rangle + \langle \partial_y \phi(I^a), \partial_y \phi(I^a) \rangle}$. It is clear that Eq.(5) is a nonlinear equation. Moreover, it is found that directly solving Eq.(5) is not feasible due to the unknown nonlinear map. In the following, we will discuss how to deal with Eq.(5). That is, one needs to apply an inner product operation to produce a kernel function. To this end, we introduce another image $I^b$ and refer to $I^b$ as a coupled image of $I^a$. Based on this, if both sides of Eq.(5) are dot-multiplied by $\phi(I^b)$, then one obtains

$$\left\langle \frac{\partial \phi(I^a)}{\partial t}, \phi(I^b) \right\rangle = \nabla \cdot (|\nabla \phi(I^a)|^{p-2} (\nabla \phi(I^a), \phi(I^b))) - \lambda(<\phi(I^a), \phi(I^b)> - <\phi(I_0^a), \phi(I^b)>), \tag{6}$$

where $(\nabla \phi(I^a), \phi(I^b)) = (<\partial_x \phi(I^a), \phi(I^b)>, <\partial_y \phi(I^a), \phi(I^b)>)$.

Similarly, we may also regard $I^a$ as the coupled image of $I^b$. Thus one can obtain another equation in terms of Eq.(6). That is, one obtains

$$\left\langle \frac{\partial \phi(I^b)}{\partial t}, \phi(I^a) \right\rangle = \nabla \cdot (|\nabla \phi(I^b)|^{p-2} (\nabla \phi(I^b), \phi(I^a))) - \lambda(<\phi(I^b), \phi(I^a)> - <\phi(I_0^b), \phi(I^a)>), \tag{7}$$

where $I_0^b$ is the degraded image of $I^b$. In such a case, we refer to Eq.(7) as the coupled equation of Eq.(6). If one adds Eq.(6) to Eq.(7), then the following equation can be obtained.

$$\frac{\partial k(I^a, I^b)}{\partial t} = \nabla \cdot (|\nabla \phi(I^a)|^{p-2} (\nabla \phi(I^a), \phi(I^b))) - 2\lambda(k(I^a, I^b) + \lambda k(I_0^a, I^b) + \lambda(k(I_0^b, I^a) +$$

$$\nabla \cdot (|\nabla \phi(I^b)|^{p-2} (\nabla \phi(I^b), \phi(I^a))). \tag{8}$$

It is observed from Eq.(8) that there is a coupled image of $I^a$. The aim of introducing the coupled image is to construct the kernel function since the implicit mapping is unknown. It is clear that the nonlinear map corresponding to the kernel function provides a new feature space. A nonlinear map corresponds to a feature space. Moreover, the kernelization technique using the coupled image may be explained as a nonlinear transformation of the original image and the nonlinear transformation may change the dynamic range of gray levels of images, thereby resulting in some image enhancement techniques in digital image processing. In other words, the total variation model based on kernel functions is actually a technique that performs the total variation in an enhanced domain of the original image. In short, the main role of kernel functions in the proposed model is to explore a proper feature space where the image is effectively processed by selecting kernel parameters.

In addition to $k$ in Eq.(8) that is a kernel function, it is found from Eq.(8) that one needs to compute the partial derivatives of mapping functions in the inner product. Fortunately, in terms of the definition of kernel functions, one can derive the following lemmas which can be used to further deal with Eq.(8).

Lemma 1:

$$<\partial_x \phi(I^a(x,y)), \phi(I^b(x,y))> = \partial_x k(I^a(x,y), I^b(x',y'))|_{x'=x, y'=y}. \tag{9}$$

Lemma 2:

$$<\partial_x \phi(I^a(x,y)), \partial_x \phi(I^b(x,y))> = \partial_{xx'} k(I^a(x,y), I^b(x',y'))|_{x'=x, y'=y}. \tag{10}$$

Lemma 3:

$$<\partial_{xx} \phi(I^a(x,y)), \phi(I^b(x,y))> = \partial_{xx} k(I^a(x,y), I^b(x',y'))|_{x'=x, y'=y}. \tag{11}$$

It is obvious that lemmas 1, 2 and 3 only give the partial derivatives of mapping functions with respect to the variable $x$. It is also straightforward to obtain the partial derivatives of mapping functions with respect to the variable $y$. From lemmas 1, 2 and 3, it is found that solving the partial derivatives of mapping functions in the inner product means computing partial derivatives of kernel functions. Specifically, the right-hand side of Eq.(8) can be computed by using kernel functions. As a result, Eq.(8) can be regarded as a kernel evolving equation in some sense and we refer to Eq.(8) as the kernel total variation model. For the sake of clarity, we also derive some partial derivatives of Gaussian kernels and polynomial kernels when dealing with color images, which are listed as follows.

If $<\phi(I^a), \phi(I^b)> = k(I^a, I^b) = \exp(-\sum_{i=1}^{3}(I_i^a - I_i^b)^2 / 2\delta)$, one obtains

$$<\partial_x \phi(I^a), \phi(I^b)> = k(I^a, I^b) \sum_{i=1}^{3}((I_i^b - I_i^a)\partial_x I_i^a / \delta), \tag{12}$$

$$<\partial_y \phi(I^a), \phi(I^b)> = k(I^a, I^b) \sum_{i=1}^{3}((I_i^b - I_i^a)\partial_y I_i^a / \delta), \tag{13}$$

$$<\partial_x \phi(I^a), \partial_x \phi(I^a)> = \sum_{i=1}^{3}(\partial_x I_i^a)^2 / \delta, \tag{14}$$

$$<\partial_y \phi(I^a), \partial_y \phi(I^a)> = \sum_{i=1}^{3}(\partial_y I_i^a)^2 / \delta, \tag{15}$$

$$<\partial_x \phi(I^a), \partial_y \phi(I^a)> = \sum_{i=1}^{3}(\partial_x I_i^a \partial_y I_i^a) / \delta. \tag{16}$$

If $<\phi(I^a), \phi(I^b)> = k(I^a, I^b) = (\sum_{i=1}^{3} I_i^a I_i^b)^d$, one obtains

$$<\partial_x \phi(I^a), \phi(I^b)> = d(\sum_{i=1}^{3} I_i^a I_i^b)^{d-1} \sum_{i=1}^{3}(I_i^b \partial_x I_i^a), \tag{17}$$

$$<\partial_y \phi(I^a), \phi(I^b)> = d(\sum_{i=1}^{3} I_i^a I_i^b)^{d-1} \sum_{i=1}^{3}(I_i^b \partial_y I_i^a), \tag{18}$$

$$<\partial_x \phi(I^a), \partial_x \phi(I^a)> = d(d-1)(\sum_{i=1}^{3} I_i^a I_i^a)^{d-2}(\sum_{i=1}^{3}(I_i^a \partial_x I_i^a))^2$$
$$+ d(\sum_{i=1}^{3} I_i^a I_i^a)^{d-1} \sum_{i=1}^{3}(\partial_x I_i^a)^2, \tag{19}$$

$$<\partial_y \phi(I^a), \partial_y \phi(I^a)> = d(d-1)(\sum_{i=1}^{3} I_i^a I_i^a)^{d-2}(\sum_{i=1}^{3}(I_i^a \partial_y I_i^a))^2$$
$$+ d(\sum_{i=1}^{3} I_i^a I_i^a)^{d-1} \sum_{i=1}^{3}(\partial_y I_i^a)^2, \tag{20}$$

$$<\partial_x \phi(I^a), \partial_y \phi(I^a)> = d(d-1)(\sum_{i=1}^{3} I_i^a I_i^a)^{d-2}(\sum_{i=1}^{3}(I_i^a \partial_x I_i^a)) \sum_{i=1}^{3}(I_i^a \partial_y I_i^a))$$
$$+ d(\sum_{i=1}^{3} I_i^a I_i^a)^{d-1} \sum_{i=1}^{3}(\partial_x I_i^a \partial_y I_i^a). \tag{21}$$

It is obvious that these equations can be derived in terms of Lemmas 1, 2, and 3. In fact, these equations can be computed in advance when applied to the proposed model.

It is found that most images we deal with belong to the function space of bounded variation(BV). The theory of bounded variation plays an important role in Eq.(1). That is, the image we try to deal with should lie in the space of bounded variation. It is obvious that we map the image into a feature space by a nonlinear map, so it makes sense to ask whether the image in the new feature space is in the space of bounded variation. Before discussing this problem, we first give the following lemma.

**Lemma 4** [21]: Let $I(x,y)$ be a function with bounded variation on a compact set and $\phi : \Re \to F$ be Lipschitz continuous on a bounded set. Then the composite map $\phi(I(x,y))$ belongs to the space of bounded variation.

Based on Lemma 4, we propose the following theorem.

**Theorem 1:** With $I(x,y)$ as before, suppose that the nonlinear mapping $\phi : \Re \to F$ in Eq.(4) corresponds to a kernel function $k(I^a(x,y), I^b(x,y))$ that is symmetric and continuous. If the function $k(I^a(x,y), I^a(x,y)) + k(I^b(x,y), I^b(x,y)) - 2k(I^a(x,y), I^b(x,y))$ satisfies the 2nd order Holder continuity, i.e.

$$k(I^a(x,y), I^a(x,y)) + k(I^b(x,y), I^b(x,y)) - 2k(I^a(x,y), I^b(x,y)) \leq L|I^a(x,y) - I^b(x,y)|^2, \quad (22)$$

where $L$ is a constant, then $\phi(I(x,y))$ is in the space of bounded variation.

Proof. $\|\phi(I^a(x,y)) - \phi(I^b(x,y))\|$

$$= \sqrt{k(I^a(x,y), I^a(x,y)) + k(I^b(x,y), I^b(x,y)) - 2k(I^a(x,y), I^b(x,y))} \quad (23)$$

If Eq.(22) holds, we substitute Eq.(22) into Eq.(23) and obtain

$$\|\phi(I^a(x,y)) - \phi(I^b(x,y))\| \leq \sqrt{L}|I^a(x,y) - I^b(x,y)|, \quad (24)$$

Equation (24) shows the map $\phi$ is Lipschitz continuous. Thus from Lemma 1, we complete the proof of Theorem 1.

Note that different nonlinear mappings may correspond to the same kernel function. Hence, in Theorem 1, we give the condition in terms of kernel functions instead of the nonlinear mapping. This shows that if we choose the kernel function in terms of Eq.(22), it is not necessary to care about the specific nonlinear map. In fact, there are numerous kernel functions in kernel-based learning. Theorem 1 only gives a sufficient condition to ensure that the mapping image belongs to the space of bounded variation, but there exists a possibility that the mapping image may be in the space of bounded variation even if Eq.(22) does not hold. According to Theorem 1, we obtain the following corollary.

**Corollary 1:** Assume that polynomial kernels $k(I^a(x,y), I^b(x,y)) = \langle I^a(x,y), I^b(x,y) \rangle^d$ and Gaussian kernels $\exp(-(|I^a(x,y) - I^b(x,y)|^2/2\delta))$ are adopted. If the kernel parameter $d$ in polynomial kernels is not smaller than 1 and the parameter $\delta$ in Gaussian kernels is bigger than zero, then the projected image $\phi(I(x,y))$ is in the space of bounded variation.

Proof: It is necessary to verify that polynomial kernels and Gaussian kernels satisfy the condition of Eq.(22). For Gaussian kernels, we have

$$k(I^a(x,y), I^a(x,y)) + k(I^b(x,y), I^b(x,y)) - 2k(I^a(x,y), I^b(x,y))$$

$$= 2 - 2\exp(-(|I^a(x,y) - I^b(x,y)|^2/2\delta)) \leq |I^a(x,y) - I^b(x,y)|^2/\delta. \quad (25)$$

Note that in Eq.(25) we use $1 - e^{-z} \leq z$ for $z \geq 0$. Thus for $\delta > 0$, Gaussian kernels satisfy Eq.(22). For polynomial kernels, we have

$$k(I^a(x,y), I^a(x,y)) + k(I^b(x,y), I^b(x,y)) - 2k(I^a(x,y), I^b(x,y))$$

$$= ((I^a(x,y))^d - (I^b(x,y))^d)^2 = [(dI^\theta(x,y))^{d-1}|I^a(x,y) - I^b(x,y)|]^2, \quad (26)$$

where $\theta$ lies in the interval of (a, b) or (b, a). In Eq.(26) we use the mean value theorem on polynomial functions. It is obvious that $[(dI^\theta(x,y))^{d-1}]^2$ is bounded if the gray level of the image takes values in a compact set and $d \geq 1$. Thus for $d \geq 1$, polynomial kernels satisfy Eq.(22). This completes the proof of corollary 1. Note that if the pixel value of images is strictly bigger than zero, the parameter *d* in polynomial kernels can be further relaxed, i.e. $d > 0$.

**3.2.2 Simplifying models for gray images**

When only considering the gray image, we can obtain the following corollary.

**Corollary 2:** If the image is a gray image and a constant image is selected as its coupled image in the case of Gaussian kernels or polynomial kernels, the Euler-Lagrange equation obtained by Eq.(8) in the kernel space is denoted by

$$-\nabla \cdot (|\nabla \phi(I^a)|^{p-2} (\nabla k(I^a, I^b))) + \lambda k(I^a, I^b) - \lambda k(I_0^a, I^b) = 0. \tag{27}$$

**Proof:** Assume that $I^b$ is a gray image and all of its pixel values are equal. For Gaussian kernels or polynomial kernels, one obtains

$$<\phi(I^a), \partial_x \phi(I^b)> = 0, \tag{28}$$

$$<\phi(I^a), \partial_y \phi(I^b)> = 0. \tag{29}$$

Note that $I^b$ is a constant gray image. It is not difficult to use Eqs.(12-16) and (17-21) to verify that Eqs.(28) and (29) hold. Further, one obtains

$$(\nabla \phi(I^a), \phi(I^b)) = (<\partial_x \phi(I^a), \phi(I^b)>, <\partial_y \phi(I^a), \phi(I^b)>), \tag{30}$$

$$\partial_x k(I^a, I^b) = <\partial_x \phi(I^a), \phi(I^b)> + <\phi(I^a), \partial_x \phi(I^b)> = <\partial_x \phi(I^a), \phi(I^b)>, \tag{31}$$

$$\partial_y k(I^a, I^b) = <\partial_y \phi(I^a), \phi(I^b)> + <\phi(I^a), \partial_y \phi(I^b)> = <\partial_y \phi(I^a), \phi(I^b)>. \tag{32}$$

From Eqs.(30), (31), and (32), one obtains

$$(\nabla \phi(I^a), \phi(I^b)) = (<\partial_x \phi(I^a), \phi(I^b)>, <\partial_y \phi(I^a), \phi(I^b)>)$$

$$= (\partial_x k(I^a, I^b), \partial_y k(I^a, I^b)). \tag{33}$$

It is also easy to verify that the following equation holds.

$$(\nabla \phi(I^b), \phi(I^a)) = (0, 0). \tag{34}$$

As $I_0^b$ is the degraded image of $I^b$, we cannot obtain $I_0^b$ in the general case. For the sake of simplicity, we generally assume that $I_0^b = I^b$. Thus substituting Eqs.(33) and (34) into Eq.(8), one can obtain the following equation.

$$-\nabla \cdot (|\nabla \phi(I^a)|^{p-2} (\nabla k(I^a, I^b))) + \lambda k(I^a, I^b) - \lambda k(I_0^a, I^b) = 0. \tag{35}$$

This completes the proof.

Note that if polynomial kernels with polynomial degree being 1 are adopted and $I^b = 1$, one can achieve Eq.(2) from Eq.(27). As a result, equation (2) is a special case of equation (27). In other words, the proposed model further generalizes the existing models in some sense.

**3.2.3 The discretization of Eq.(27)**

It is of interest to note that Eq.(27) involves computing the partial derivatives of kernel functions instead of partial derivatives of the original image and the coupled image. Hence one can directly discretize kernel functions in terms of partial derivatives of kernel functions as it is convenient to deal with kernel functions. We also follow the numerical scheme in Ref. (22) as the gradient descent method is usually slow due to the time step constraints imposed by numerical stability. Assume that the current location is $O=(x,y)$. Let $E=(x,y+1)$, $N=(x,y-1)$, $S=(x,y+1)$, $W=(x-1,y)$ denote its four neighbors, and $e=(x,y+0.5)$, $n=(x,y-0.5)$, $s=(x,y+0.5)$, $w=(x-0.5,y)$ be the four midway points. Note that we cannot directly obtain the kernel value of these four midway points. Let $v=(v^1,v^2)=|\nabla\phi(I^a)|^{p-2}(\nabla k(I^a,I^b))$. Then the divergence is first discretized by using the central difference, denoted by

$$\nabla \cdot v = \frac{\partial v^1}{\partial x} + \frac{\partial v^2}{\partial y} \approx v_e^1 - v_w^1 + v_n^2 - v_s^2, \tag{36}$$

where $v_e^1 \approx |\nabla\phi(I^a)|^{p-2}(k(I^a,I^b)_E - k(I^a,I^b)_O)$, $v_w^1 \approx |\nabla\phi(I^a)|^{p-2}(k(I^a,I^b)_W - k(I^a,I^b)_O)$, $v_n^2 \approx |\nabla\phi(I^a)|^{p-2}(k(I^a,I^b)_N - k(I^a,I^b)_O)$, and $v_s^2 \approx |\nabla\phi(I^a)|^{p-2}(k(I^a,I^b)_S - k(I^a,I^b)_O)$. It is noted that $|\nabla\phi(I^a)|^{p-2} = (\langle\partial_x\phi(I^a),\partial_x\phi(I^a)\rangle + \langle\partial_y\phi(I^a),\partial_y\phi(I^a)\rangle)^{\frac{p-2}{2}}$. $\langle\partial_x\phi(I^a),\partial_x\phi(I^a)\rangle$ and $\langle\partial_y\phi(I^a),\partial_y\phi(I^a)\rangle$ can be computed from the above lemmas. Thus from Eqs.(12-16) and (17-21), one can find that these inner products can be obtained in terms of gradient information and pixel values. Let $\Lambda=\{E,N,W,S\}$. Equation (27) can be discretized as follows.

$$0 = \sum_{R\in\Lambda} |\nabla\phi(I^a)|^{p-2}(k(I^a,I^b)_O - k(I^a,I^b)_R) + \lambda(k(I^a,I^b)_O - k^0(I_0^a,I^b)_O) \tag{37}$$

Let us define the following equations.

$$w_R = |\nabla\phi(I^a)|^{p-2}, \tag{38}$$

$$h_R = \frac{w_R}{\sum_{P\in\Lambda} w_P + \lambda}, \tag{39}$$

$$h_O = \frac{\lambda}{\sum_{P\in\Lambda} w_P + \lambda}, \tag{40}$$

It is not difficult to verify that $\sum_{R\in\Lambda} h_R + h_O = 1$. Hence Eq.(37) can also be expressed as follows.

$$k(I^a,I^b)_O = \sum_R h_R k(I^a,I^b)_R + h_O k(I_0^a,I^b)_O. \tag{41}$$

It is found that Eq.(41) has the form of a low pass filter, which is a system of nonlinear equations as the coefficients of this filter depend on kernel functions. Adopting the Gauss-Jacobi iteration scheme for linear systems at each step $t$, one can update the kernel function $k^{(t-1)}$ to $k^{(t)}$ by

$$k^{(t)}(I^a, I^b)_O = \sum_R h_R^{(t-1)} k^{(t-1)}(I^a, I^b)_R + h_O^{(t-1)} k^{(t-1)}(I_0^a, I^b)_O. \tag{42}$$

As $h$ in Eq.(42) is actually a low pass filter [22], the iterative process from Eq.(42) is stable.

### 3.2.4 Obtaining processed images

It is found that the proposed model is performed in the kernel function space due to the fact that introducing a coupled image is to form kernel functions. Thus from Eq.(8) or Eq.(42), one can only obtain the updated value of kernel functions. It is still necessary to use the processed image in the image space to update the value of kernel functions. In other words, in each iteration we need to obtain the processed image in the image space from the updated value of kernel functions. This can be done by considering the characteristics of kernel functions. In order to get the processed image in the image space in an effective way, the coupled images should not be set randomly. From the viewpoint of practical applications, one should adopt different coupled images for three channels of color images. In the following we show how to obtain the processed image by setting the coupled image of color images in the case of Gaussian kernels and polynomial kernels.

Assume that we have already obtained the processed image at the $(t-1)th$ iteration, denoted by $I^{a^{(t-1)}}(x, y) = (I_1^{a^{(t-1)}}(x, y), I_2^{a^{(t-1)}}(x, y), I_3^{a^{(t-1)}}(x, y))$. For Gaussian kernels, in order to obtain the processed result of the first component (channel), we first choose another color image as the coupled image of $I^{a^{(t-1)}}(x, y)$. This coupled image is denoted by $I^b(x, y)$ and its three components satisfy $I_1^b(x, y) = 0, I_2^b(x, y) = I_2^{a^{(t-1)}}(x, y)$, and $I_3^b(x, y) = I_3^{a^{(t-1)}}(x, y)$. The reason for choosing such a coupled image is that the component corresponding to the first channel can be obtained easily. By doing so, we construct the kernel function from $I^{a^{(t-1)}}(x, y)$ and $I^b(x, y)$, denoted by

$$k(I^{a^{(t-1)}}(x, y), I^b(x, y)) = \exp(-(I_1^{a^{(t-1)}}(x, y))^2 / 2\delta). \tag{43}$$

Using $I^{a^{(t-1)}}(x, y)$, $I^b(x, y)$ and the kernel function in Eq.(43), we can obtain the updated value of the kernel function from Eq.(8), denoted by $k_g^1$. After obtaining this value of the kernel function, we can get the first component of the color image by the following equation.

$$I_1^{a^{(t)}}(x, y) = \sqrt{-\ln(k_g^1) \cdot 2\delta}. \tag{44}$$

Thus the first component of the color image is updated. Similarly, in order to update the second component of color images, we first choose a coupled image such that $I_1^b(x, y) = I_1^{a^{(t)}}(x, y)$, $I_2^b(x, y) = 0$ and $I_3^b(x, y) = I_3^{a^{(t-1)}}(x, y)$. Then we construct the kernel function from this coupled image and obtain the updated value of the kernel function from Eq.(8). The second component of the color image can be obtained by using a similar formula of Eq.(44), denoted by $I_2^{a^{(t)}}(x, y)$. For the third component of color images, the coupled image should satisfy $I_1^b(x, y) = I_1^{a^{(t)}}(x, y), I_2^b(x, y) = I_2^{a^{(t)}}(x, y)$, and $I_3^b(x, y) = 0$. The updated value of the third component of color images is also obtained in a similar way. By doing so, all of the three components of the color image are updated. Obviously, each channel of color images corresponds to a coupled image in such a case.

In the case of polynomial kernels, in order to obtain the processed result of the first component $I_1^{a^{(t-1)}}(x,y)$, one may select $I^b(x,y)$ such that $I_1^b(x,y)=1$, $I_2^b(x,y)=0$ and $I_3^b(x,y)=0$. Then we construct the kernel function from $I^{a^{(t-1)}}(x,y)$ and $I^b(x,y)$, denoted by

$$k(I^{a^{(t-1)}}(x,y), I^b(x,y)) = (I_1^{a^{(t-1)}}(x,y))^d . \qquad (45)$$

Using $I^{a^{(t-1)}}(x,y)$, $I^b(x,y)$ and the kernel function in Eq.(45), we can obtain the updated value of the kernel function from Eq.(8), denoted by $k_p^1$. After that, the first component of the color image can be updated by the following equation.

$$I_1^{a^{(t)}}(x,y) = \sqrt[d]{k_p^1} . \qquad (46)$$

In a similar way, other two components of the color image can be updated. Note that we give how to deal with color images using two types of kernel functions. In fact, for gray images, it is only necessary to deal with one component and the updated value of kernel functions can be obtained by Eq.(42). Overall, in order to obtain the processed image in the image space, one needs to implement Eq.(44) or Eq.(46). It is obvious that we easily obtain the image in the kernel function space without using Eq.(44) or Eq.(46).

## 4  Experimental Results and Discussion

In this section, we evaluate the performance of the proposed model on gray and color images.

### 4.1 The selection of kernel parameters

As the proposed model involves kernel functions, the parameters in kernel functions will directly affect the performance of the proposed model. It is necessary to determine the optimal kernel parameter in the proposed model. In this subsection, we will discuss how to select proper kernel parameters in order to achieve good performance. Note that an image can be regarded as a two-dimensional surface in a Hilbert space in the general case. The area of a surface in the feature space can be computed by using the kernel trick without knowing implicitly mapping. That is, the area of this surface can be computed by using $A(\bar{X}) = \iint \sqrt{g}\, dxdy$ [23], where $g$ is a metric tensor. As a result, different kernel parameters will give different areas of a surface. Let us define a ratio by $r = \dfrac{A(\bar{X})}{A(X)}$, where $A(\bar{X})$ is the area of the surface in the mapping space and $A(X)$ is the area of the surface in the original space. This ratio can be explained as an indicator of the variation on the surface. If $r=1$, it shows that the mapping can preserve areas.

In this set of experiments, we check the ratio $r$ with different kernel parameters on the Lena image with $256 \times 256$ pixels. The corrupted images are obtained by adding multiplicative Gaussian noise with zero mean and standard deviation (20) to the original image. In this paper we consider two types of kernel functions (Gaussian kernels and polynomial kernels) due to the fact that they satisfy the condition of Eq.(22). In order to reduce the staircase effect, the parameter $p$ in Eq.(8) is set to 1.2 in this paper. The polynomial degree $d$ in polynomial kernels varies from 1.1 to 2 by an interval of 0.1, and the kernel parameter $\delta$ is in Gaussian kernels varies from 0.1 to 1 by an interval of 0.1. We also use the PSNR to evaluate the quality of processed images. Figure 1 (a) shows the PSNR of the proposed model and the ratio as polynomial kernel parameters vary in the case of the double Y axes on the Lena image, and Figure 1 (b) shows the PSNR of our model and the ratio with the change of Gaussian kernel parameters on the Lena image.

As can be seen from Figure 1(a), the ratio $r$ becomes big with the increase of the kernel parameter $d$. However, the PSNR of this image sees a downward trend as the kernel parameter $d$ increases. It is found that the PSNR of this image is the highest when the ratio $r$ is close to one. So it is better to select the optimal kernel parameter if the ratio $r$ lies 1 or so. From Figure 1(b), it is found that the ratio $r$ decreases as the kernel parameter increases. It is observed that the PSNR of this image first increases significantly and then becomes relatively steady. This shows that we should choose the optimal kernel parameters when the ratio $r$ is in the vicinity of 1. Overall, these experimental results show that by properly selecting the kernel parameters in two types of kernels, one can obtain good PSNR of images.

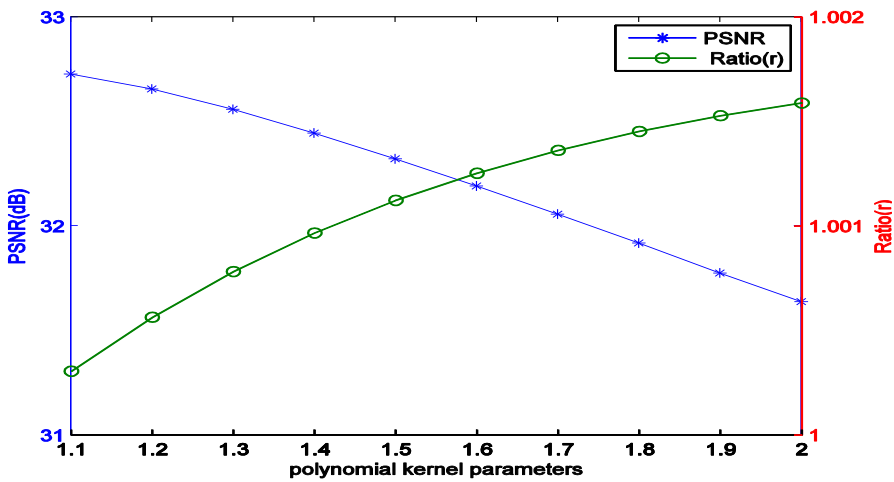

(a) polynomial kernels

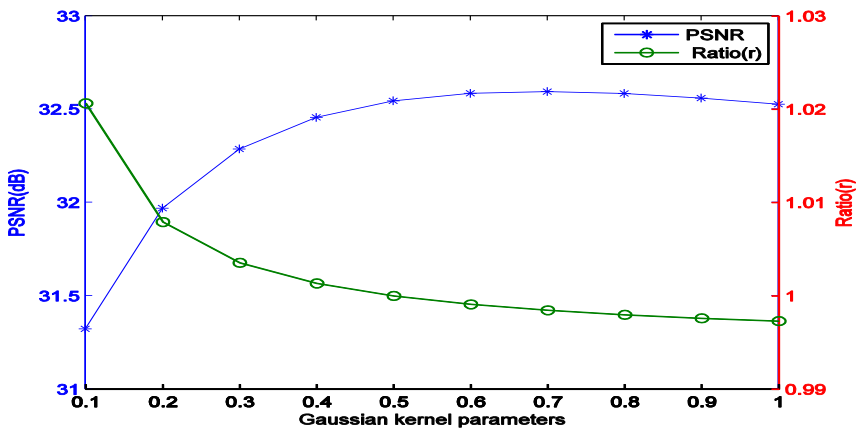

(b) Gaussian kernels

Figure 1: PSNR of the proposed model and the ratio with varying kernel parameters

It is found that we perform the proposed model in the kernel function space. When the image is embedded into the kernel function space, it makes sense to ask what change has happened. As shown in Figure 3 (b), Figure 3 (c), Figure 4(b), and Figure 4 (c), the visual quality of images has been changed when they are embedded into the kernel function space. In order to assess the visual quality of images in the kernel function space, it is necessary to use image quality assessment models. Although there are numerous models in previous literature[24,25], we adopt the natural image quality evaluator (NIQE)[24] in this paper due to its simplicity. In the NIQE model, the NIQE index is the performance index for evaluating the image quality. In the general case, the bigger the NIQE index is, the worse the visual quality of images. Figure 2 shows the experimental results on the Lena image. From Figure 2 (a), it is found that if images are embedded into the polynomial kernel space, the NIQE index will change as kernel parameters vary. As pointed out in Section 3.2.2, when $d=1$ and $I^b(x,y)=1$, our model degenerate to the TV model, so in such a case the NIQE index is actually computed from the noisy image in the image space. This shows that by choosing proper kernel parameters, the image in the kernel space has better visual quality than the noisy image in the image space. Likewise, we find from Figure 2(b) that the NIQE index of the image in the Gaussian kernel space may be lower than that of the noisy image by choosing kernel parameters. This shows again that embedding the image into the kernel function space may improve the visual quality of images in some sense. This also helps us explain why our method may obtain good performance in some cases and this simultaneously provides a strategy to choose kernel parameters in the proposed model by using NIQE. However, note that NIQE is used to evaluate natural images in real world. The lowest NIQE index may not mean the highest PSNR since NIQE does not requires reference images and PSNR makes use of reference images. In the following section, when we apply the proposed model to denoise images, we use PSNR as performance measure; when we enhance images by the proposed model, we adopt NIQE as the performance measure.

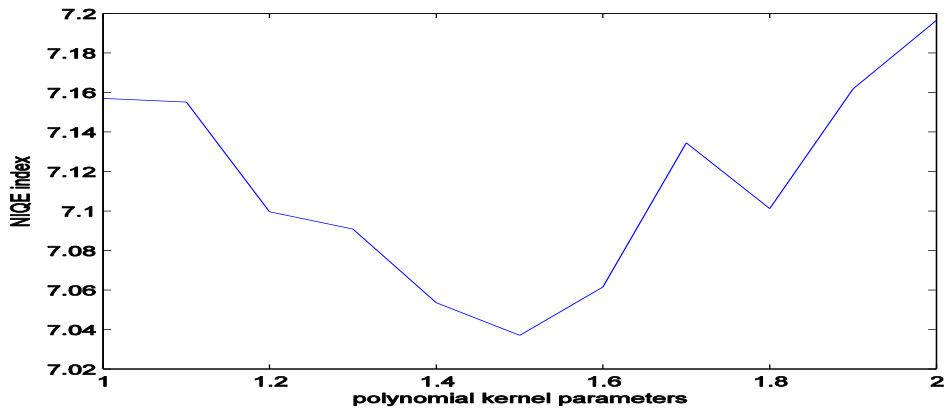

(a) polynomial kernels

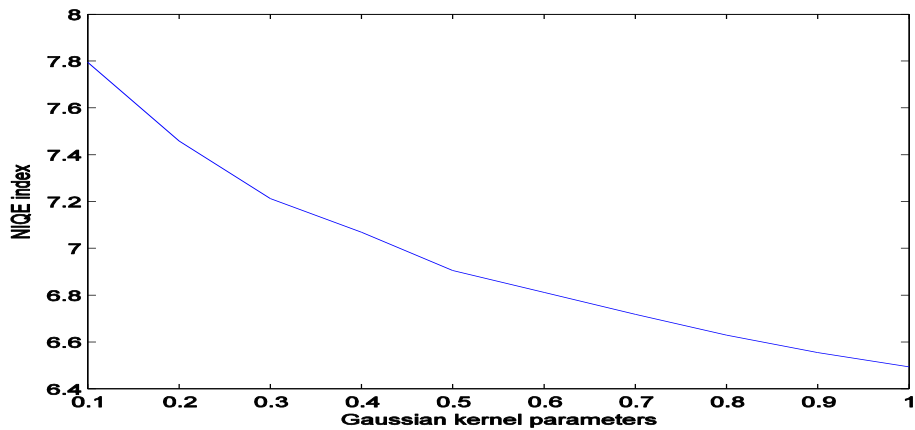

(b) Gaussian kernels

Figure 2: NIQE index of the image in the kernel space with varying kernel parameters

## 4.2 Denoising Gray Images

In this set of experiments, we evaluate the proposed model on Lena, House, Couple, Hill, Boat, Barbara, and Peppers images. We also test the proposed model in terms of Gaussian kernels(GK) and polynomial kernels(PK), denoted by GK+GTV and PK+GTV respectively. Note that the nonlocal total variation model (NLTV) which utilizes more information from images can also improve the performance of the TV model. It is not difficult to embed kernel functions into the NLTV model by using the basic idea of this paper. We also implement NLTV [16-17] in terms of Gaussian kernels and polynomial kernels, denoted by GK+NLTV and PK+NLTV respectively although we do not give the detailed derivation of this model. In implementing NLTV-based methods, we use ten best neighbors in the searching window of $5 \times 5$ and the 4 nearest neighbors to compute the weight. Currently there are some sparse representation methods that can be used to denoise images [26-29]. It is noted that the sparse denoising method based on the adaptive dictionary produced by the K-SVD algorithm leads to the state-of-the-art performance. However, this method requires estimating the variance of noise in the general case. We actually find that this method is consistently superior to the proposed model in dealing with additive Gaussian noise if the variance of noise is known, but the variance of noise is unknown and the types of noise are diverse. To this end, in this paper, we also test the performance of the sparse denoising method from K-SVD in dealing with multiplicative Gaussian noise.

In the case of multiplicative Gaussian noise, we test the proposed model on noisy images with a relatively small deviation (20) and a relatively large deviation (80). The maximal number of iterations of the algorithms is set to be 50, and the tradeoff parameter $\lambda$ is set to be 10 in the case of the small deviation and 1 in the case of the large deviation. We also choose kernel parameters based on the idea of Section 4.1. Figure 3 shows the some experimental results on the Lena image. Figure 3(a) shows the noisy image. Figure 3(b) shows the image obtained by performing a nonlinear transformation $\langle I^a(x,y), I^b(x,y) \rangle^d$ with $d$=1.1 and $I^b(x,y)=1$. Figure 3(c) shows the negative of the image obtained by performing a nonlinear transformation $\exp(-(|I^a(x,y) - I^b(x,y)|^2 / 2\delta))$ with $I^b(x,y)=0$ and $\delta = 0.5$. Note that the Gaussian function $\exp(-|I^a(x,y)|^2)$ does not preserve the increasing order from black to white in the output image, so we show $1 - \exp(-|I^a(x,y)|^2)$ to reserve the order from black to white. We also show the denoised images and the residual images of the Lena image in terms of the best PSNR in low noise in Fig.3(d)–(j2).

From Figure 3, it seems that NLTV-related models obtain higher brightness than GTV-related ones. It is obvious that embedding kernel functions into existing total variation models can further improve the denoising performance of images. Table 1 also lists the PSNR of several methods for dealing with seven images and the best performance of each image in two noisy cases is highlighted in bold. From Table 1, one can see that the PSNR of processed images decreases with the increase of standard deviations. When the standard deviation is 20, it is found that PK+NLTV obtains the best performance of all the methods on House, Hill, Boat, Barbara and Peppers images, and GK+NLTV performs best on Lena and Couple images. When the standard deviation is 80, PK+NLTV is better than other methods on House, Hill, Barbara Boat and Peppers images, and GK+NLTV is the best on Lena and Couple images. It is noted that the classical GTV model is worse than PK+GTV or GK+GTV in the general case. One can see that combining kernel functions into NLTV also contributes to the improvements of performance. That is, PK+NLTV or GK+NLTV can obtain higher PSNR than NLTV. It is also found that there is no clear winner for applying Gaussian kernels or polynomial kernels on these images. A possible explanation is that the kernels we use may depend on the characteristics of noisy images. It is noted that NLTV gives better performance than GTV since the former considers more information from images. It is also found that K-SVD does not show a clear superiority in dealing with multiplicative Gaussian noise, especially for high noise, but this method is still better than GTV or NLTV in the general case. Overall, embedding kernel functions into total variation models can improve the filtering results for noisy images. As a result, one can embed Gaussian kernels or polynomial kernels into total variation models to deal with the images with multiplicative noise.

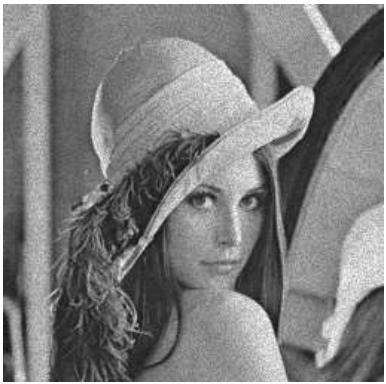
(a)

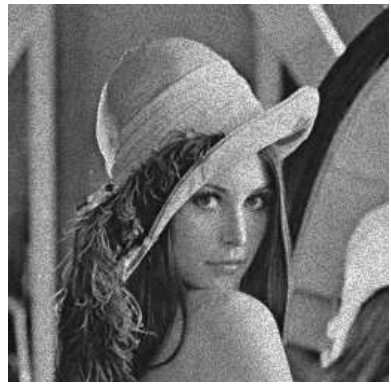
(b)

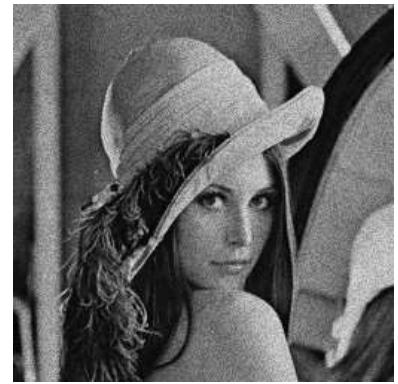
(c)

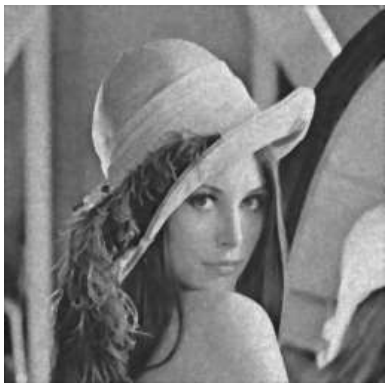
(d1)

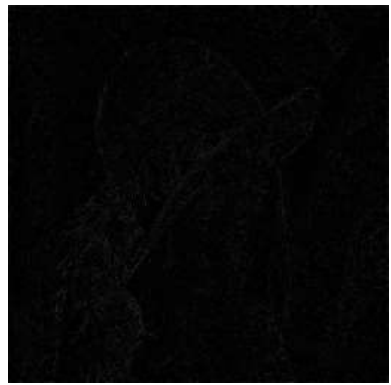
(d2)

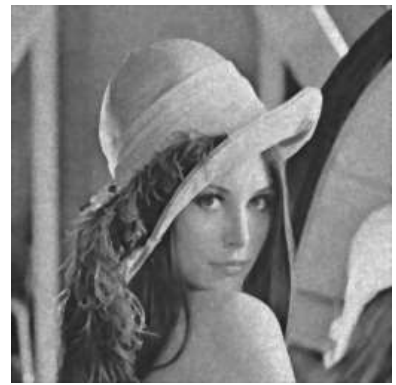
(e1)

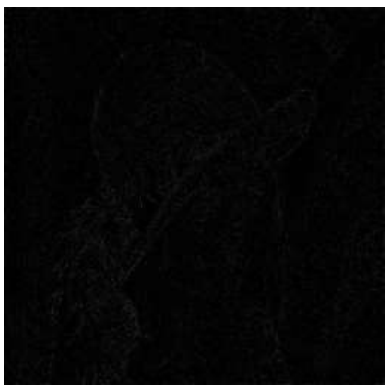
(e2)

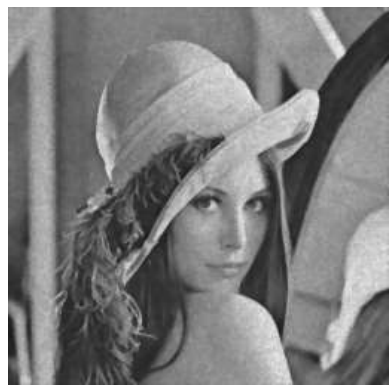
(f1)

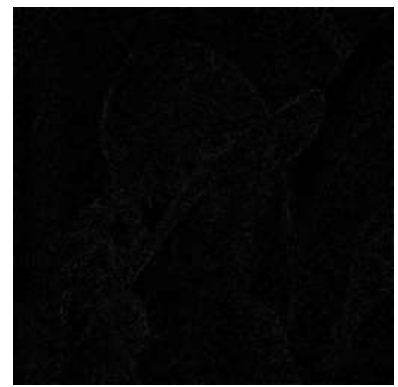
(f2)

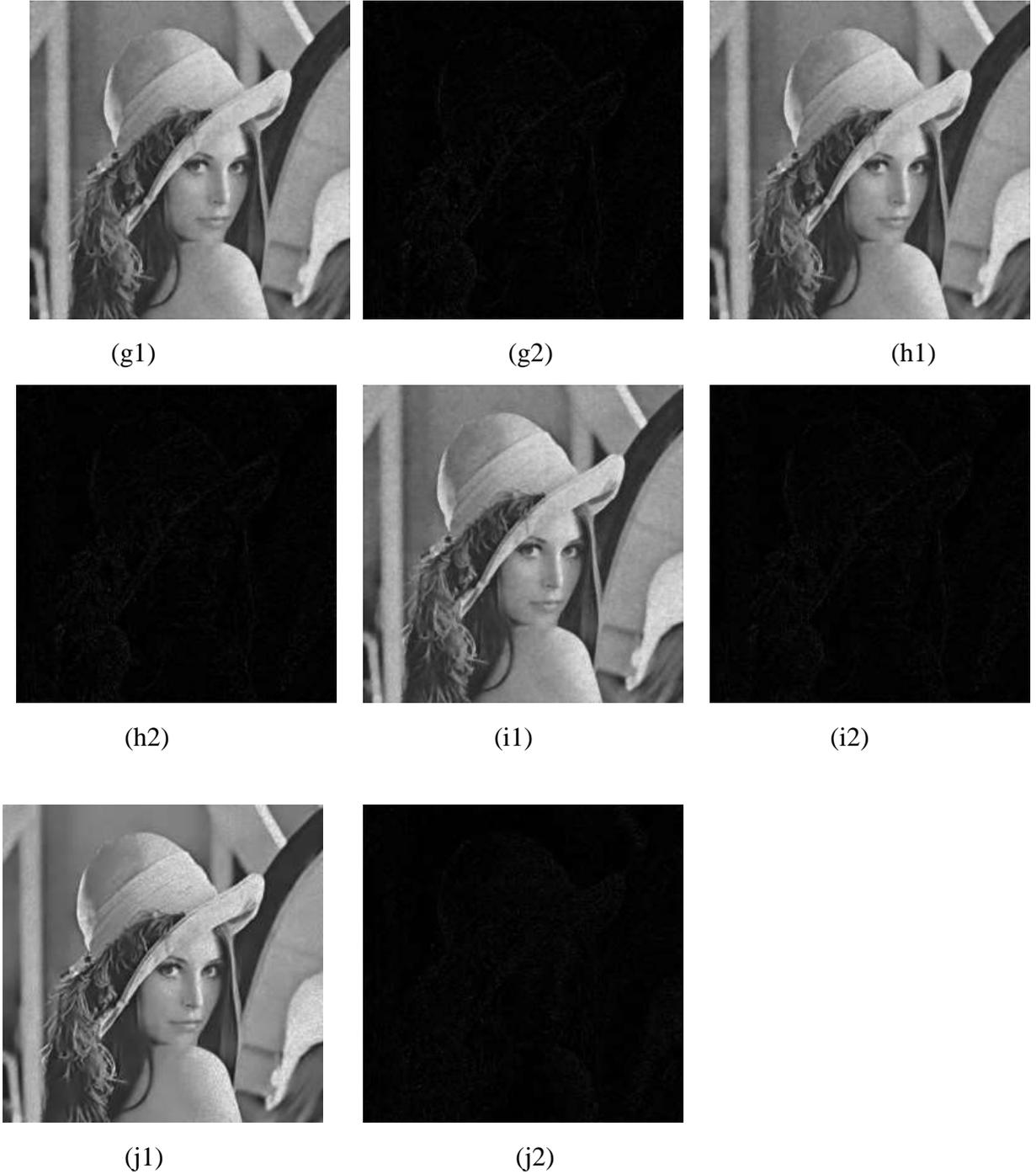

Figure 3: the experimental results on the Lena image (a) the noisy Lena image; (b) the image obtained in the kernel function space $\left\langle I^a(x,y), I^b(x,y) \right\rangle^d$ with $d=1.1$; (c) the negative of the image obtained in the kernel function space $\exp(-(\left|I^a(x,y)-I^b(x,y)\right|^2/2\delta))$ with $\delta=0.5$; (d1) the denoising image based on GTV; (d2) the absolute value of the residual image obtained by GTV; (e1) the denoising image based on PK+GTV; (e2) the absolute value of the residual image obtained by PK+GTV; (f1) the denoising image based on GK+GTV; (f2) the absolute value of the residual image obtained by GK+GTV; (g1) the denoising image based on NLTV; (g2) the absolute value of the residual image obtained by NLTV; (h1) the denoising image based on PK+NLTV; (h2) the absolute value of the residual image obtained by PK+NLTV; (i1) the denoising image based on GK+NLTV; (i2) the absolute value of the residual image obtained by GK+NLTV; (j1) the denoising

image based on K-SVD; (j2) the absolute value of the residual image obtained by K-SVD;

Table 1: PSNR(dB) of models for handling gray images

|    | images  | Lena  | House | Hill  | Couple | Boat  | Barbara | Peppers |
|----|---------|-------|-------|-------|--------|-------|---------|---------|
| 20 | GTV     | 32.16 | 32.71 | 32.02 | 31.88  | 30.82 | 30.80   | 31.76   |
|    | GK+GTV  | 32.45 | 32.94 | 32.23 | 32.09  | 30.93 | 30.89   | 32.12   |
|    | PK+GTV  | 32.21 | 33.08 | 32.27 | 32.01  | 31.06 | 31.12   | 32.15   |
|    | NLTV    | 32.54 | 32.94 | 32.35 | 31.91  | 31.05 | 30.90   | 31.89   |
|    | GK+NLTV | **32.99** | 33.15 | 32.51 | **32.17** | 31.43 | 31.41 | 32.23 |
|    | PK+NLTV | 32.87 | **33.21** | **32.65** | 31.98 | **31.59** | **31.52** | **32.42** |
|    | K-SVD   | 32.88 | 32.98 | 31.78 | 32.13  | 31.37 | 31.33   | 32.03   |
| 80 | GTV     | 24.92 | 24.98 | 24.74 | 23.88  | 23.68 | 24.46   | 24.36   |
|    | GK+GTV  | 25.24 | 25.82 | 25.33 | 24.12  | 23.99 | 24.77   | 24.72   |
|    | PK+GTV  | 24.39 | 25.92 | 25.43 | 24.17  | 23.96 | 24.90   | 24.77   |
|    | NLTV    | 25.26 | 26.02 | 25.02 | 23.93  | 23.95 | 24.79   | 25.16   |
|    | GK+NLTV | **25.53** | 26.42 | 25.48 | **24.26** | 24.12 | 24.98 | 25.45 |
|    | PK+NLTV | 25.38 | **26.50** | **25.64** | 24.08 | **24.32** | **25.05** | **25.59** |
|    | K-SVD   | 24.90 | 25.12 | 24.30 | 23.95  | 23.97 | 23.93   | 24.00   |

**4.3. Denoising Color images**

In this subsection, we continue to verify the proposed model on color images whose sizes are $256 \times 256 \times 3$ pixels. The selection of kernel parameters and other settings are similar to the strategies in Section 4.2.

In this set of experiments, we carry out the experiments on some color images including House, Lena, F16, and Peppers images. The corrupted images are obtained by adding multiplicative Gaussian noise with zero mean and two different standard deviations (20 and 80) to the original images. Figure 4 shows the experimental results on the House image. From Figure 4, it is found that combining kernel functions into GTV or NLTV can further enhance the performance of the corresponding model. The PSNR of each method is shown in Table 2. From Table 2, PK+NLTV performs best on the Lena, House and Peppers images, and GK+NLTV obtains the best performance on the F16 images. One can see that the model using Gaussian kernels or polynomial kernels is generally superior to the corresponding one. Different from the results on gray images, the improvements on color images is relatively obvious. This is partly because the information from three channels in color images is effectively combined by using kernel functions. Like the results on the gray images, the model using polynomial kernels is not consistently better than the one using Gaussian kernels and NLTV is superior to GTV in the general case. It is found that PK+GTV or GK+GTV is better than GTV, and NLTV is worse than PK+NLTV or GK+NLTV. We also note that K-SVD is better than GTV in most cases and is worse than PK+NLTV or GK+NLTV. Overall, these experimental results show again that it is reasonable to embed Gaussian kernels or polynomial kernels into the existing total variation models to deal with color images with multiplicative noise.

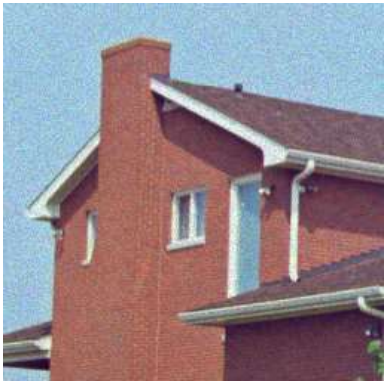 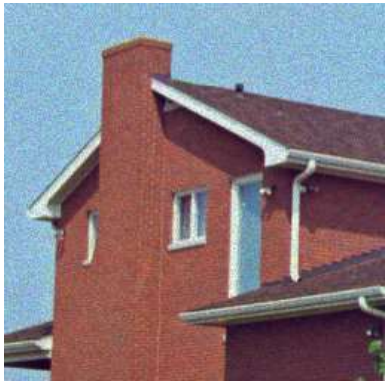 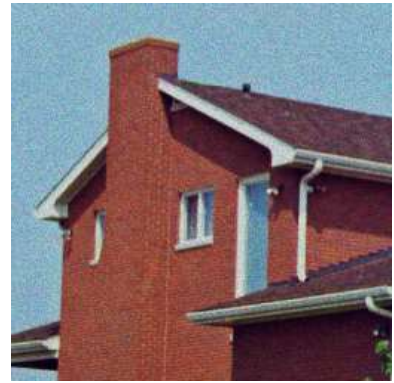

(a)　　　　　　　　　　　　(b)　　　　　　　　　　　　(c)

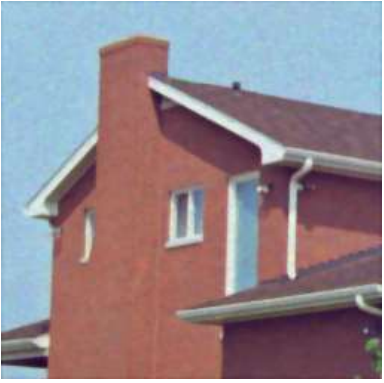 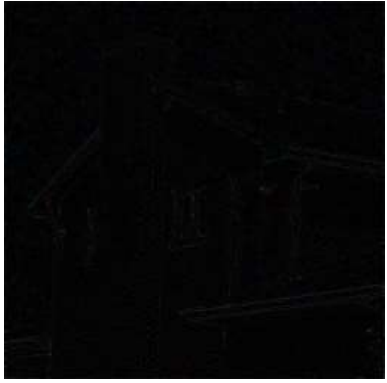 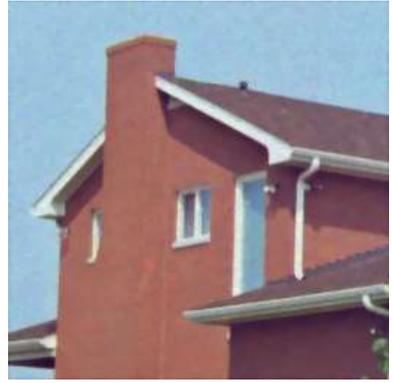

(d1)　　　　　　　　　(d2)　　　　　　　　　(e1)

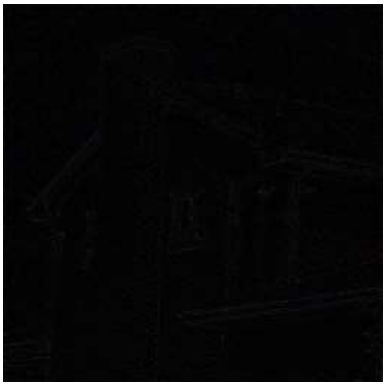 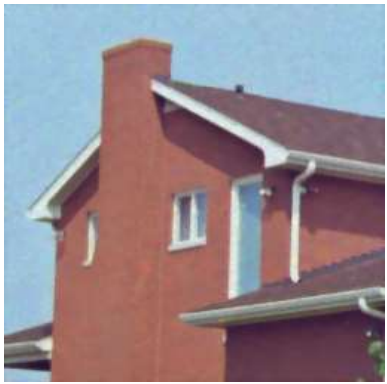 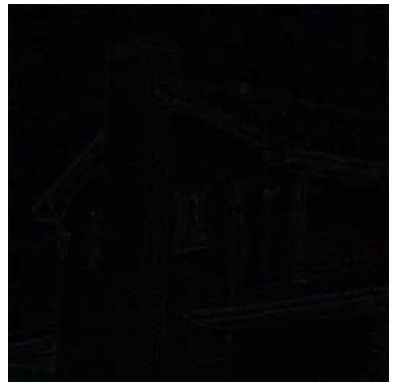

(e2)　　　　　　　　　(f1)　　　　　　　　　(f2)

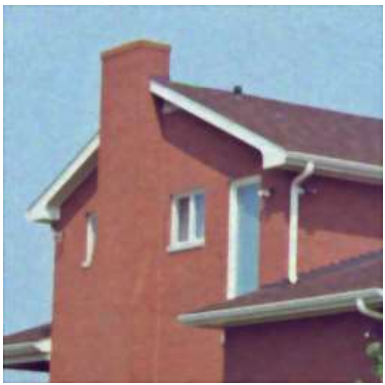 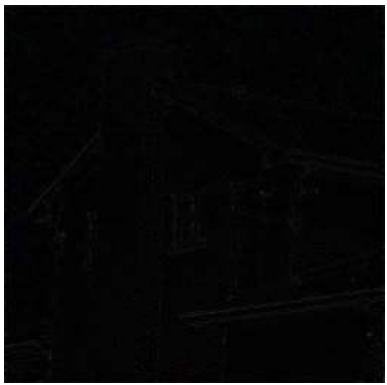 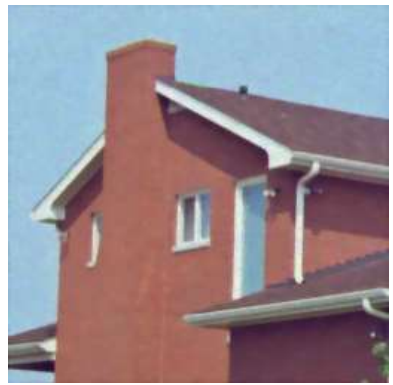

(g1)　　　　　　　　　(g2)　　　　　　　　　(h1)

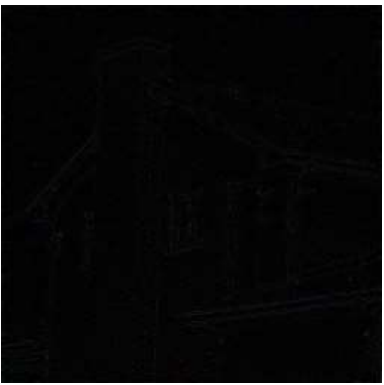 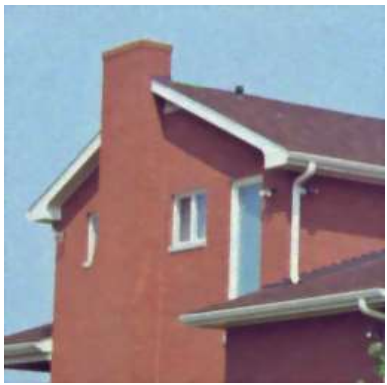 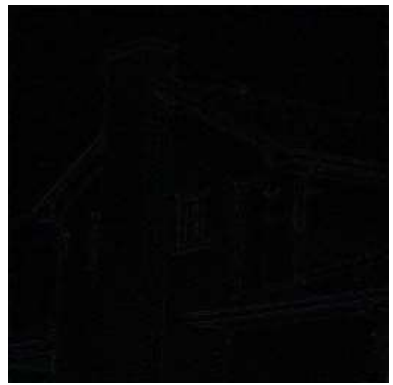

(h2)　　　　　　　　　(i1)　　　　　　　　　(i2)

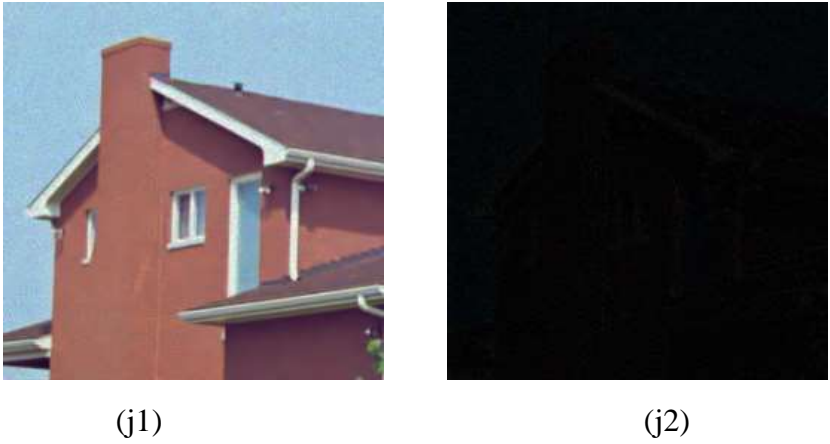

(j1)            (j2)

Figure 4: the experimental results on the House image (a) the noisy house image; (b) the image obtained in the kernel function space $\langle I^a(x,y), I^b(x,y) \rangle^d$ with d=1.1; (c) the negative of the image obtained in the kernel function space $\exp(-(|I^a(x,y) - I^b(x,y)|^2 / 2\delta))$ with $\delta = 0.5$; (d1) the denoising image based on GTV; (d2) the absolute value of the residual image obtained by GTV; (e1) the denoising image based on PK+GTV; (e2) the absolute value of the residual image obtained by PK+GTV; (f1) the denoising image based on GK+GTV; (f2) the absolute value of the residual image obtained by GK+GTV; (g1) the denoising image based on NLTV; (g2) the absolute value of the residual image obtained by NLTV; (h1) the denoising image based on PK+NLTV; (h2) the absolute value of the residual image obtained by PK+NLTV; (i1) the denoising image based on GK+NLTV; (i2) the absolute value of the residual image obtained by GK+NLTV; (j1) the denoising image based on K-SVD; (j2) the absolute value of the residual image obtained by K-SVD;

Table 2: PSNR(dB) of denoising models for handling color images

| Methods | | GTV | GK+GTV | PK+GTV | NLTV | GK+NLTV | PK+NLTV | K-SVD |
|---|---|---|---|---|---|---|---|---|
| Lena | 20 | 31.94 | 32.55 | 32.67 | 32.07 | 32.89 | **32.95** | 32.09 |
|  | 80 | 24.38 | 24.77 | 24.93 | 24.73 | 25.30 | **25.36** | 24.81 |
| House | 20 | 32.44 | 33.09 | 33.16 | 32.87 | 33.32 | **33.47** | 32.77 |
|  | 80 | 24.67 | 25.51 | 25.48 | 25.11 | 25.87 | **25.97** | 24.42 |
| F16 | 20 | 30.21 | 31.20 | 30.93 | 30.53 | **31.34** | 31.09 | 30.33 |
|  | 80 | 22.57 | 23.03 | 23.17 | 22.99 | **23.78** | 23.41 | 22.87 |
| Peppers | 20 | 32.51 | 33.22 | 32.88 | 32.70 | 33.34 | **33.48** | 32.69 |
|  | 80 | 25.10 | 25.51 | 25.17 | 25.63 | 25.79 | **25.94** | 25.09 |

## 4.4 Automatic Contrast Enhancement

In this subsection, we explore the proposed model for automatic image enhancement. It is noted that classical contrast enhancement is often achieved by applying linear or nonlinear transformation on an original image so as to change the gray level of images. There are several basic types of functions widely used for image enhancement including linear, logarithmic and power-law transformation. It is found that in the proposed method the image we process is in the kernel function space. In fact these kernel functions respond to nonlinear transformation in image enhancement. Instead of obtaining the original pixel value from Eq.(46) or Eq.(44), we directly display the image in the kernel function space, thereby enhancing the quality of the original image. It is obvious that different kernel parameters will give different images. It is necessary to select the best image from those that have been produced. Fortunately, some state-of-the art 'general purpose' no reference image quality assessment (IQA) algorithms can be used to predict the quality of output images from different kernel parameters. Note that these models are suitable for assessing some natural images. As pointed out in Section 4.1, it may be unsteady in evaluating those images used in Section 4.2. To this end, we test our proposed model on natural images. It is also noted that in Section 4.2 and 4.3 NLTV, PK+NLTV, and GK+NLTV are better than GTV, PK+GTV and GK+NTV respectively, so we only test the former three methods in this subsection. In order to choose the best quality of images, the Nature Image Quality Evaluator (NIQE) index [24] is used in this paper due to its simplicity. Figure 5 shows the experimental results on a natural image. Note that we deal with gray images and color images independently instead of obtaining the result of gray images from the result of color images. Figure 5(a) is the original gray image whose NIQE index is 5.98. By using NLTV to deal with the image in Figure 5(a), we obtain the image in Figure 5(b) whose NIQE index is 4.72. Figure 5(c) is the image obtained by using PK+NLTV, but it belongs to the polynomial kernel function space and the kernel parameter is determined by the NIQE index. That is, the resulting image is not handled by Eq.(46). Figure 4(d) is the image obtained by GK+NLTV and we display the negative of the image since adopting Gaussian kernels makes the gray level of an image reverse. In a similar way, we can obtain those images in Figure 5(e), (f), (g), and (h). As shown in Figure 5, for gray images, the image obtained by GK+NLTV has the lowest NIQE index, so it has the best visual quality in these four gray images; for color images, GK+NLTV obtains the best visual quality of images. Therefore, from the perspective of image enhancement, handling images in the kernel function space corresponds to performing some operations in the enhanced domain of images, which helps us build a bridge between TV models and some image enhancement techniques by kernel functions. As a result, if the images are shown in the proper kernel function space, we actually enhance the image in some sense.

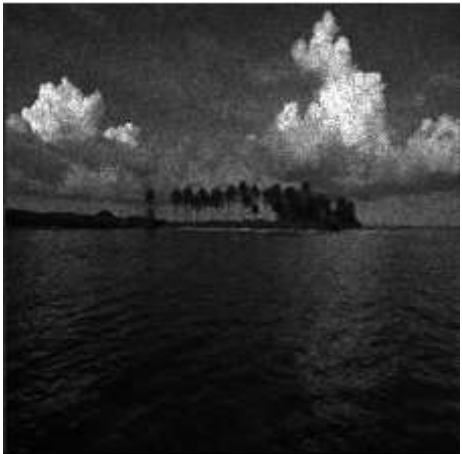
(a)  NIQE index=5.98

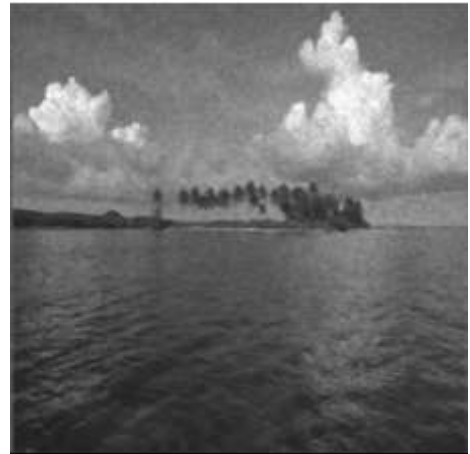
(b) NIQE index=4.92

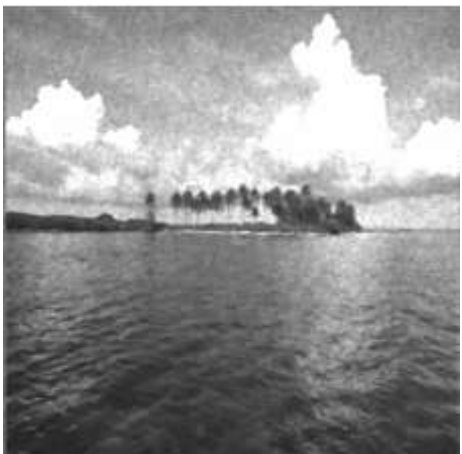
(c) NIQE index=4.64

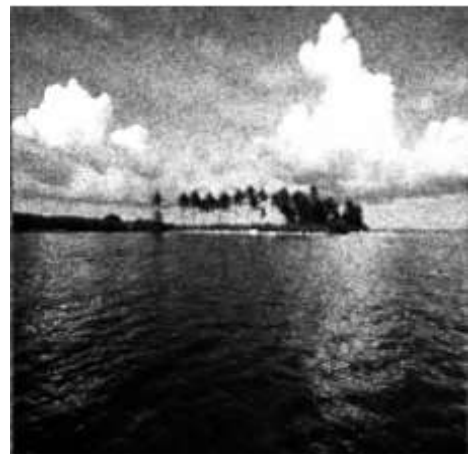
(d)  NIQE index=2.93

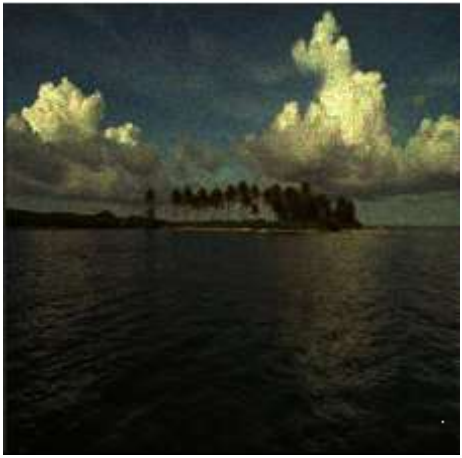
(e) NIQE index=17.23

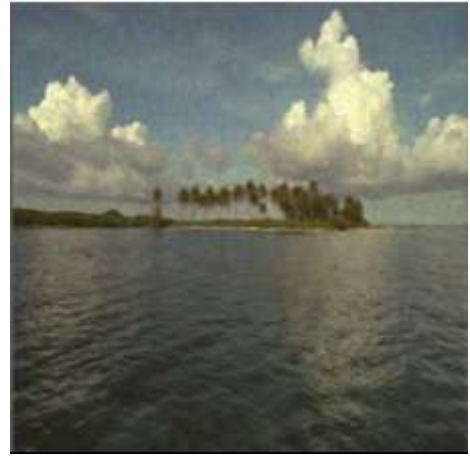
(f) NIQE index=13.28

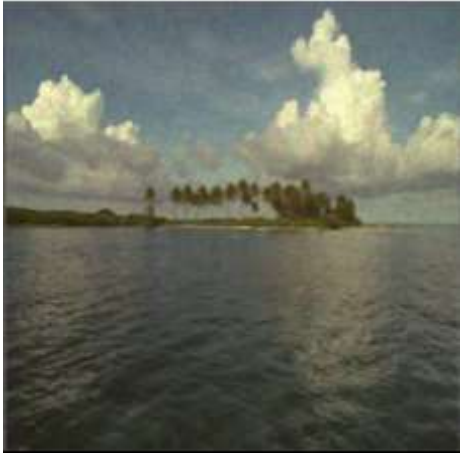
(g) NIQE index=12.19

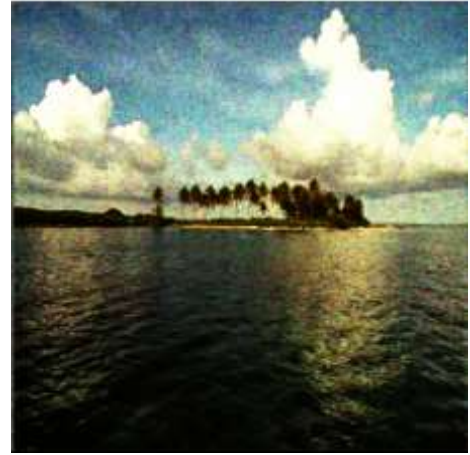
(h) NIQE index=11.37

**Figure 5:** Experimental results on a natural image; (a) the original gray image (b) the gray image processed by NLTV; (c) the enhanced gray image in the polynomial kernel space from PK+NLTV ; (d) the enhanced gray image in the Gaussian kernel space from GK+ NLTV; (e) the original color image; (f) the color image processed by NLTV; (g) the enhanced color image in the polynomial kernel space from PK+NLTV; (h) the enhanced color image in the Gaussian kernel space from GK+ NLTV

## 5 Conclusions

This paper proposes a generalized TV model based on kernel functions to deal with gray or color images. The proposed model incorporates kernel methods into the generalized total variation model to enhance its performance. We give a sufficient condition to ensure that the embedding space is in the space of bounded variation. Via our theoretical analysis, it is found that the proposed model further extends the generalized total variation model. In some sense, the proposed method can be regarded as a technique that performs the generalized total variation on an enhanced domain of images, which connects TV models and some image enhancement techniques. As experimentally demonstrated, the proposed model gives better performance than previous models in some cases, which shows that our model can benefit from kernel functions. But there is still much work to be done. For example, it will be appealing if kernel parameters in the proposed model are chosen locally instead of globally. In addition, it should be pointed out that the basic idea of this paper can be applied to other partial differential nonlinear equations. We expect to generalize this idea to other nonlinear equations in the near future.

## 6 Acknowledgments:

This work is partially supported by the FRF for the Central Universities (2015XKMS084).

# 7 References

1. F.Tokamani-Azar and K.E.Tait, "Image recovery using the anisotropic diffusion equation," IEEE Trans. Pattern Anal. Machine Intell., pp.1573-1578 (1996)
2. L.Alvarez, P.Lions, and J. Morel, "Image selective smoothing and edge detection by nonlinear diffusion," SIAM J. Numer.Anal., 29(3), pp.845-866 (1992)
3. M.Zhu, S.J.Wright, and T.F.Chan, "Duality-based algorithms for total variation image restoration,"Compute.Optim. Appl., pp.08-33(2008)
4. L.Rudin, S.Osher, and E.Fatemi, "Nonlinear total variation based noise removal algorithms," Physica D, 60, pp.259-268(1992)
5. B.Song, "Topics in variational PDE image segmentation, inpainting and denoising," Ph.D. Thesis, USA, University of California Los Angeles, June, (2003)
6. S.Osher, M.Burger, D.Goldfarb, J.Xu, and W.Yin, "An iterative regularization method for total variation based on image restoration," Multiscale modeling and Simulation, 4(2) pp.460-489(2005)
7. S.Esedoglu and S.Osher, "Decomposition of images by the anisotropic Rudin-Osher-Fatemi model," Communications in Pure and Applied Mathematics, 57(12), pp.1609-1626(2004)
8. K.Bredies, K.Kunisch, and T.Pock, "Total generalized variation," SIAM Journal on Imaging Sciences, vol. 3, no. 3, pp. 492-526(2010)
9. F. Knoll, K.Bredies, T.Pock, and R.Stollberger, "Second order total generalized variation (TGV) for MRI," Magnetic Resonance in Medicine, vol. 65, no. 2, pp. 480-491(2011)
10. D. M.Strong and T. F.Chan, "Edge-preserving and scale dependent properties of total variation regularization," Inverse Problems, vol. 19, no. 6, pp. S165-S187(2003)
11. Q.Chen, P.Montesinos, Q. S.Sun, P. A.Heng and D. S. Xia, "Adaptive total variation denoising based on difference curvature," Image and Vision Computing, vol. 28, no. 3, pp. 298-306(2010)
12. T. F.Chan and S.Esedoglu, "Aspects of total variation regularized L1 function approximation," SIAM Journal on Applied Mathematics, vol. 65, no. 5, pp.1817-1837 (2005)
13. T.Chan, S.Esedoglu, F. Park, and A.Yip "Recent developments in total variation image restoration," Mathematical Models of Computer Vision, Springer, New York, NY, USA, 2005
14. M.Bertalm and S. Levine, "Denoising an image by denoising its curvature image," SIAM Journal on Imaging Sciences, vol. 7, no. 1, pp.187-211(2014)
15. T.Chan, A.Marquina, and P. Mulet, "High-order total variation-based image restoration," SIAM Journal on Scientific Computing, vol. 22, no. 2, pp.503-516(2000)
16. G.Gilboa and S. Osher, "Nonlocal operators with applications to image processing," Multiscale Model. Simul., vol. 7, no. 3, pp.1005-1028(2008)
17. Y. Lou, X.Zhang, S.Osher, and A.Bertozzi, "Image recovery via nonlocal operators," J. Sci. Comput., vol. 42, no. 2, pp.185-197(2010)
18. B.Scholkopf, A.Smola, and K.R. Miiler, "Nonlinear component analysis as a kernel eigenvalue problem," Neural computation, 10, pp.1299-1319(1998)
19. B.Sholkopf and A.J.Smola, Learning with kernels, MIT Press, Cambridge, MA, (2002)
20. Z.Liang, Y.Li, and T.Zhao, "Projected gradient method for kernel discriminant nonnegative matrix factorization and the applications," Signal Processing, vol.90, pp.2150-2163(2010)
21. J.J.MoreAu and M.VaLadier, "A chain rule involving vector function of bounded variation," Journal of Functional analysis, 74, pp.222-345(1987)
22. T.F.Chan and J.Shen, "Mathematical models for local non-texture inpainting," SIAM J. Appl.

Math., 62(3), pp.1019-1043(2002)
23. G.Aubert and P.Kornprobst, Mathematical problems in image processing: partial differential equations and the calculus of variations, Springer, Applied mathematical Sciences, vol.147, (2006)
24. A.Mittal, R.Soundarajan, and A. C.Bovik, "Making a 'Completely Blind' Image Quality Analyzer," IEEE Signal Processing Letters , vol. 20, no. 3, pp. 209-212( 2013)
25. M.Saad, A. C.Bovik, and C.Charrier, "Blind image quality assessment: A natural scene statistics approach in the DCT domain," IEEE Trans. Image Process., vol. 21, no. 8, pp. 3339-3352(2012)
26. M.Elad and M. Aharon, "Image denoising via sparse and redundant representations over learned dictionaries," IEEE Trans. Image Process., vol. 15, no. 12, pp. 3736-3745(2006)
27. M.Protter and M.Elad, "Image sequence denoising via sparse and redundant representations," IEEE Trans. Image Process., vol. 18, no. 1, pp. 27-35(2009)
28. O. G Guleryuz, "Weighted averaging for denoising with overcom-plete dictionaries," IEEE Trans. Image Process., vol. 16, no. 12, pp. 3020-3034(2007)
29. S. Li, H. Yin, and L.Fang, "Group-sparse representation with dictionary learning for medical image denoising and fusion', IEEE Trans. Biomed. Eng., vol. 59, no. 12, pp. 3450–3459(2012)